\documentclass[runningheads]{llncs}
\usepackage[T1]{fontenc}
\newcommand{\repeatthanks}{\textsuperscript{\thefootnote}}

\usepackage{graphicx}
\usepackage{xcolor}
\usepackage{amsmath}
\usepackage{hyperref}
\usepackage{lipsum}

\begin{document}
%

\title{Image2Text2Image: A Novel Framework for Label-Free Evaluation of Image-to-Text Generation with Text-to-Image Diffusion Models}
\titlerunning{Image2Text2Image: A Novel Framework for Label-Free Evaluation} 

\newcommand{\blfootnote}[1]{%
  \begingroup
  \renewcommand\thefootnote{}\footnote{#1}
  \addtocounter{footnote}{-1}
  \endgroup
}
%
%


\linespread{0.95}





\author{
Jia-Hong Huang\thanks{Equal contribution} \and Hongyi Zhu\repeatthanks \and Yixian Shen \and Stevan Rudinac  \and Evangelos Kanoulas}

\authorrunning{J. Huang et al.}
\institute{University of Amsterdam, Amsterdam, The Netherlands}


%
\maketitle              

\begin{abstract}
Evaluating the quality of automatically generated image descriptions is a complex task that requires metrics capturing various dimensions, such as grammaticality, coverage, accuracy, and truthfulness. 
Although human evaluation provides valuable insights, its cost and time-consuming nature pose limitations.
Existing automated metrics like BLEU, ROUGE, METEOR, and CIDEr attempt to fill this gap, but they often exhibit weak correlations with human judgment. To address this challenge, we propose a novel evaluation framework called Image2Text2Image, which leverages diffusion models, such as Stable Diffusion or DALL-E, for text-to-image generation.
In the Image2Text2Image framework, an input image is first processed by a selected image captioning model, chosen for evaluation, to generate a textual description. Using this generated description, a diffusion model then creates a new image. By comparing features extracted from the original and generated images, we measure their similarity using a designated similarity metric. A high similarity score suggests that the model has produced a faithful textual description, while a low score highlights discrepancies, revealing potential weaknesses in the model's performance. Notably, our framework does not rely on human-annotated reference captions, making it a valuable tool for assessing image captioning models.
Extensive experiments and human evaluations validate the efficacy of our proposed Image2Text2Image evaluation framework. The code and dataset will be published to support further research in the community.


\keywords{Image Captioning \and Metrics for Automated Evaluation \and Text to Image Generation Models}

\end{abstract}

\section{Introduction}

Evaluating sentences generated by automated methods remains a significant challenge in image captioning. Existing metrics for assessing image descriptions aim to capture various desirable qualities, including grammaticality, coverage, correctness, truthfulness, and more. While human evaluation is crucial for accurately quantifying these attributes—often using Likert scales or pairwise comparisons \cite{mitchell2012midge,rohrbach2013translating,zhang2024comparative,huang2024multi,wang2024prototype}—it is expensive, difficult to replicate, and time-consuming. This has led to an increasing demand for automated evaluation measures that closely align with human judgment. The primary challenge in developing such metrics lies in integrating these diverse evaluation criteria into a cohesive measure of sentence quality.

Several automated metrics, such as BLEU \cite{papineni2002bleu}, ROUGE \cite{lin2004rouge}, METEOR \cite{elliott2013image}, and CIDEr \cite{vedantam2015cider}, have been developed to evaluate image descriptions generated by automated methods. BLEU, originally designed for machine translation, focuses on precision, while ROUGE, from the summarization domain, emphasizes recall. METEOR aims to assess the overall quality of image descriptions. However, studies have shown that these metrics often exhibit a weak correlation with human judgment \cite{bybaby,elliott2013image,rudinac2013learning,zhang2024beyond}.
In contrast, the consensus-based metric CIDEr measures the similarity between a generated sentence and a set of human-authored ground truth sentences, and it generally aligns well with human consensus. However, CIDEr's effectiveness depends on having a sufficiently large and diverse set of ground truth sentences, which can be a limitation when such data is scarce \cite{vedantam2015cider}. This limitation also applies to other methods like CLAIR \cite{chan2023clair} and the previously mentioned metrics. Additionally, some approaches involve caption ranking \cite{hodosh2013framing}, but they often fall short in evaluating novel image descriptions.


To address the challenge of evaluating image descriptions, we introduce a novel framework that leverages modern vision large language models (VLLMs), such as GPT-4 \cite{brown2020language} or Gemini \cite{team2023gemini}, capable of creating images. The advancements in VLLMs, exemplified by models like GPT-4V, enable the creation of textual prompts that generate images closely aligned with the semantic meaning of the input text.
The core design philosophy of our proposed framework is that if an image captioning model is effective, its generated description should be accurate enough to reconstruct the original image, or a highly similar one, using VLLM-based image generation. The continuous evolution of VLLM technology underpins and strengthens the foundation of this framework.


Our proposed framework for evaluating image captioning models starts by defining the task of taking an image as input and generating a corresponding textual description. The input image is processed by the image captioning model under evaluation, producing a descriptive text. This description is then fed into a text-to-image generation model, such as Stable Diffusion, to generate a new image based on the text. Next, we extract features from both the original input image and the generated image, comparing them using the cosine similarity metric. 
Importantly, our framework does not require human-annotated reference captions for this evaluation. In this framework, a high cosine similarity score indicates that the generated description is of high quality, allowing the text-to-image generation to accurately recreate an image that closely resembles the original. Conversely, a low cosine similarity score suggests that the text description lacks accuracy, leading to a generated image that diverges from the original. This discrepancy signals the suboptimal performance of the image captioning model. Moreover, we apply our framework to detect multi-modal hallucinations in VLLMs, particularly in VQA and image description responses. After extensive experiments, our method shows a significant advantage in detecting hallucinations generated by VLLMs compared to other baseline models.  
Consequently, the proposed framework proves valuable for assessing the effectiveness of a given image captioning model.


\noindent\vspace{+3pt}
The key contributions of this work can be summarized as follows:
\begin{itemize}
    \item \textbf{Novel Framework for Evaluating Image-to-Text Models:} We present a novel framework that uses the text-to-image generation model, such as Stable Diffusion or DALL-E, to evaluate the quality of image descriptions generated by an image captioning model. The proposed evaluation framework does not necessitate human-annotated reference captions.

    \item \textbf{Human Assessment of the Framework:} To verify the effectiveness of our evaluation framework and facilitate large-scale comparison with human judgment, we introduce a new dataset by modifying and enhancing well-known benchmarks, originally made for conventional computer vision tasks.

    \item  \textbf{Comprehensive Experiments on Established Datasets:} We perform extensive experiments to demonstrate the efficacy of the proposed evaluation framework using widely used image captioning datasets. We also test the capability of our evaluation framework to make the human likert scale judgment and detect the hallucinations contained in the VLLM-generated image captions. 
\end{itemize}
\section{Related Work}
In this section, we begin by reviewing existing related literature, covering topics such as the image captioning methods and the evolution of automated metrics.
\vspace{-0.2cm}
\subsection{Image Captioning Methods}
The encoder-decoder network architecture has become fundamental in image captioning, as demonstrated by various studies \cite{xiao2019deep,vinyals2015show,zhu2023video,huang2017vqabq}. Typically, these networks use a CNN as the encoder to extract global image features and an RNN as the decoder to generate word sequences. Mao et al. \cite{mao2016generation} introduce a method for generating referring expressions, which describe specific objects or regions within an image. In \cite{wang2016image}, a bidirectional LSTM-based method for image captioning leverages both past and future information to learn long-term visual-language interactions.
Attention mechanisms have significantly improved the performance of image captioning models. Pedersoli et al. \cite{pedersoli2017areas} introduce an area-based attention model that predicts the next word and the corresponding image regions at each RNN timestep. While these advancements are significant, they mainly focus on single-image-based description generation. However, certain abstract concepts or descriptions might not be fully captured using only image data \cite{laserson2018textray}. \cite{huang2022non,huang2021deepopht} have explored using expert-defined keyword sequences to enhance model capabilities in generating more accurate and contextually relevant descriptions.
Recent advancements have also explored transformer-based architectures, such as Vision Transformers (ViT), which have shown promise in capturing finer details and global context in images for caption generation \cite{dosovitskiy2020image}. Furthermore, integrating multimodal learning approaches, where models are trained on both visual and textual data, has led to significant improvements in generating contextually richer and more nuanced image descriptions \cite{lu2019vilbert}.

The domain of medical image captioning has seen significant advancements, particularly through methods that combine human expertise with algorithmic capabilities. \cite{li2018hybrid} developed a Hybrid Retrieval-Generation Reinforced Agent, which integrates human prior knowledge with AI-based caption generation for medical images. This agent alternates between a generative module and a retrieval mechanism that uses a template database reflecting human expertise, producing multi-faceted, sequential sentences. \cite{jing2017automatic} contributed to this field with a multi-task learning framework that simultaneously predicts tags and generates captions. Their method, which focuses on abnormal areas in chest radiology images using an attention mechanism and a hierarchical LSTM, offers detailed descriptions.
These methods primarily focus on generating reports for chest radiology images, which differ in object size and detail compared to retinal images \cite{laserson2018textray,huang2021deepopht}. Additionally, the color features in chest radiology and retinal images differ significantly, with the former being predominantly grey-scale and the latter being colorful \cite{laserson2018textray,huang2021deepopht}. Most existing methods rely primarily on the image input for caption generation.
Recent advancements also include enhancing the CNN-RNN framework with the TransFuser model \cite{huang2022non}. This model adeptly combines features from different modalities and addresses the challenge of incorporating unordered keyword sequences with visual inputs, minimizing information loss \cite{huang2022non}. This development represents a significant stride in medical image captioning, reflecting the growing complexity and capability of these methods.
Further progress in deep learning, particularly the application of ViTs, has offered promising results in medical imaging \cite{chen2021vit}. ViTs excel in capturing intricate details and providing a broader context for more accurate medical image analysis and caption generation.
The evaluation framework proposed in this paper is versatile and capable of assessing any existing image captioning approaches. 

\subsection{Automatic Metrics for Image Captioning}
The evolution of image captioning has been significantly influenced by the development and application of automatic metrics for evaluating caption quality \cite{papineni2002bleu,lin2004rouge,banerjee2005meteor,vedantam2015cider,khan2024exquisitor,huang2024novel}. These metrics guide the training of captioning models and provide a scalable means for performance assessment. The BLEU score, a pioneering metric by \cite{papineni2002bleu}, measures n-gram precision in the generated text against a reference. ROUGE \cite{lin2004rouge} emphasizes recall through the overlap of n-grams and longest common subsequences.
Subsequent innovations introduced refined approaches. METEOR \cite{banerjee2005meteor} aligns more closely with human judgment by incorporating synonym matching and stemming. In \cite{vedantam2015cider}, the CIDEr metric, specifically designed for image captioning, assesses the similarity of generated captions to a set of reference captions. The SPICE metric \cite{anderson2016spice} evaluates semantic content and the depiction of objects, attributes, and relationships. Additionally, the NLG-Eval toolkit~\cite{sharma2017relevance} provides a comprehensive suite of metrics for a more holistic evaluation of natural language generation.
However, these metrics have limitations. Metrics like BLEU and ROUGE often fail to capture the contextual nuances of captions \cite{papineni2002bleu,lin2004rouge}. The challenge of evaluating creativity and novelty in caption generation is also evident, as automated metrics may penalize deviations from standard references \cite{vedantam2015cider,anderson2016spice}. Recently, advancements like BERTScore \cite{zhang2019bertscore} and CLIPScore \cite{hessel2021clipscore}, which utilize contextual embeddings and visual-textual alignment, respectively, have been proposed to address these challenges.

In this study, human evaluation is employed to validate the effectiveness of the proposed evaluation framework.

\begin{figure*}[ht!]
  \centering  
  \vspace{-0.5cm}
  \scalebox{0.9}{
  \includegraphics[width=\textwidth]{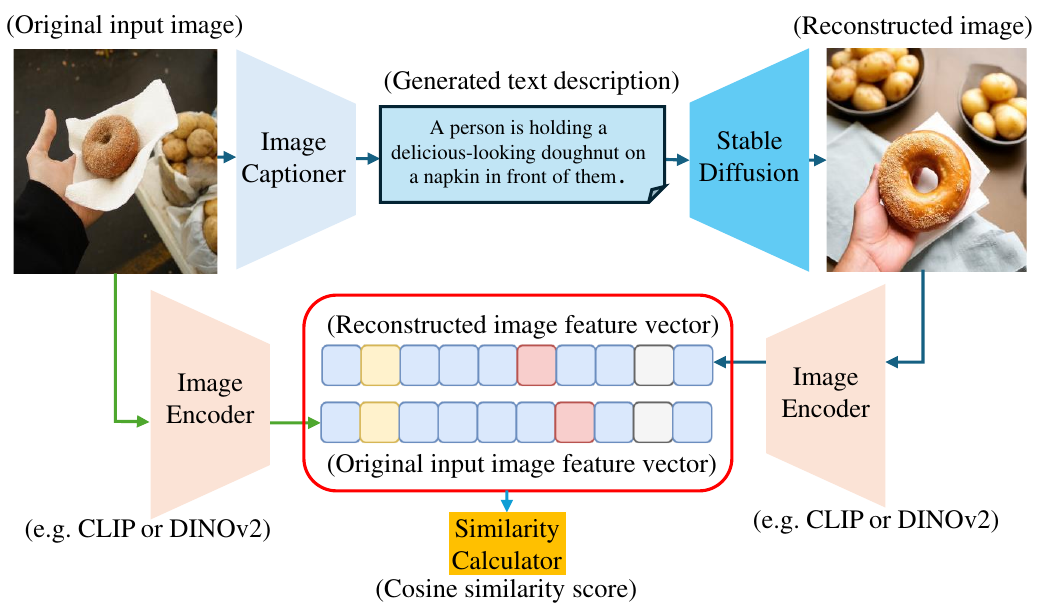}
  }
  \vspace{-0.5cm}
  \caption{
  Flowchart of the proposed evaluation framework. The proposed framework consists of four main components: an image captioning module, an image encoder, a text-to-image generation model (Stable Diffusion), and a similarity calculator. The image captioning module employs a chosen model to process an input image and generate textual descriptions. The image encoder is tasked with extracting features from the input image. The text-to-image generation model utilizes the text descriptions produced by the image captioning model to generate the corresponding image. Finally, the similarity calculator computes the similarity between the features of the input image and the image generated by the text-to-image generation model.} 
  \label{fig:figure2}
  \vspace{-1.0cm}
\end{figure*}

\section{Methodology}
The evaluation framework consists of four main components: an image captioning module, a Stable Diffusion-based text-to-image generator, an image feature extraction module, and a similarity calculator, as illustrated in Figure \ref{fig:figure2}. The following subsections will provide a detailed introduction to each component. 
\vspace{-0.2cm}
\subsection{Image Captioning Module}
The module includes an image captioning model that will be evaluated using the proposed framework. This model generates a text description from an input image. To help users understand the evaluation framework, we use the InstructBLIP model \cite{dai2305instructblip} as an example in Section~\ref{exp:Experiments and Analysis}. This example demonstrates the entire process of using the framework to evaluate an image captioning model, making it clear and accessible for users.

\subsection{Stable Diffusion-based Text-to-Image Generator}
Numerous studies \cite{brown2020language,team2023gemini,zhang2024qfmts,zhang2024towards,huang2024optimizing,Li2021Paint4PoemAD} have shown that text-to-image generators, such as GPT-4V and Stable Diffusion, can produce high-quality images that closely match the semantic meaning of given text prompts. Within the proposed framework, the Diffusion model-based image generator uses the text description generated by the preceding image captioning model. If the image captioning model performs well and produces an accurate and high-quality description, the Diffusion model-based image generator will create an image similar to the original input image. This demonstrates the connection between effective image captioning and the generation of corresponding images by the Diffusion model-based approach.

\vspace{-0.1cm}
\subsection{Image Feature Extraction Module}
\vspace{-0.1cm}
The image feature extraction module primarily consists of a pre-trained image encoder. This module takes an image as input and produces a feature vector representing the input image. To enhance user understanding of the proposed evaluation framework, we use DINOv2 \cite{oquab2023dinov2} as an example for image feature extraction in Section \ref{exp:Experiments and Analysis}. DINOv2 is a vision-only self-supervised model achieved by pre-training a ViT \cite{dosovitskiy2020image} on 142 million images without labels or annotations. Through this pre-training process, the model is distilled into smaller models with millions of parameters, achieving notable performance across various vision downstream tasks, including image recognition, video action recognition, object detection, instance segmentation, and semantic segmentation, all without extensive supervised fine-tuning. The demonstration in Section \ref{exp:Experiments and Analysis} highlights the complete process, including image feature extraction for calculating similarity scores between the input and generated images. It illustrates how the proposed framework can be used to assess an image captioning model, providing users with a clear understanding. It is worth noting that the image feature extractor can be substituted with other pre-trained CNNs, such as VGG-16 \cite{simonyan2014very} or ResNet-52 \cite{he2016deep}. Unlike vision language models such as CLIP \cite{Radford2021LearningTV} and Coca \cite{yu2022coca}, which pre-train the image encoder by aligning text-image pairs, self-supervised vision-only models focus solely on vision information extraction, making them more suitable for image-to-image comparison.
\vspace{-0.3cm}

\subsection{Similarity Calculator}
\label{metric:cosine}
\vspace{-0.2cm}
Cosine similarity, as defined in Equation~(\ref{eq:cossim}), serves as a metric for quantifying the similarity between two vectors in a multi-dimensional space. It evaluates the cosine of the angle between these vectors, offering insight into their degree of similarity or dissimilarity.
The advantage of cosine similarity lies in its ability to assess directional similarity rather than magnitude, rendering it robust against variations in scale and orientation. This characteristic makes it a widely adopted metric in diverse domains, including image processing and NLP. In these fields, cosine similarity is frequently employed to assess the similarity between images, documents, or sentences represented as vectors in high-dimensional spaces. The cosine similarity value $\text{CosSim}(\cdot~, \cdot) \in [-1, 1]$, where a value of $1$ signifies that the vectors are identical, $0$ indicates orthogonality (i.e., no similarity), and $-1$ indicates complete dissimilarity or opposition.
\begin{equation}
    \text{CosSim}(\mathbf{i_{o}}, \mathbf{i_{g}}) = \frac{\mathbf{i_{o}} \cdot \mathbf{i_{g}}}{\|\mathbf{i_{o}}\| \|\mathbf{i_{g}}\|},
\label{eq:cossim}
\end{equation}
where $\mathbf{i_{o}} \cdot \mathbf{i_{g}}$ denotes the dot product (also known as the inner product) of the original input image feature vector $\mathbf{i_{o}}$ and the LLM-generated image feature vector $\mathbf{i_{g}}$. $\|\mathbf{i_{o}}\|$ and $\|\mathbf{i_{g}}\|$ represent the Euclidean norms (also known as the magnitudes or lengths) of vectors $\mathbf{i_{o}}$ and $\mathbf{i_{g}}$, respectively. In other words, cosine similarity measures the cosine of the angle between two vectors, which represents their similarity in direction and magnitude. 
\vspace{-0.2cm}

\section{Experiments and Analysis}
\label{exp:Experiments and Analysis}
\vspace{-0.2cm}
In this section, we aim to evaluate the effectiveness of the proposed evaluation framework for image captioning models. To do this, we will validate our framework using our newly introduced image caption dataset, widely adopted human-annotated image captioning judgment datasets, and multi-modal hallucination detection datasets.
For the image captioning datasets, since all datasets have undergone human annotation, our primary objective is to determine whether the evaluation results from our framework align with human consensus or judgment. Specifically, a correct caption—matching the human-annotated counterpart—should yield a high cosine similarity score between the generated and original images, as measured by our framework. Conversely, an incorrect caption—deviating from the human-annotated version—should result in a lower cosine similarity score.
For the hallucination detection datasets, we aim to test whether our evaluation framework assigns a higher score to the accurate image description compared to an image caption containing textual descriptions not present in the image.
This approach allows us to empirically validate the effectiveness of our proposed evaluation framework in aligning with human judgment and detecting patterns of hallucinations produced by the VLLMs.
\vspace{-0.3cm}
\subsection{Experimental Settings}
To demonstrate the application of the proposed framework for evaluating an image captioning model, we use the InstructBLIP \cite{dai2305instructblip} model in our image captioning module. This model incorporates the pre-trained language model Vicuna-7B \cite{Zheng2023JudgingLW} to generate image descriptions. Image captions are generated using the prompt ``<Image> A short image caption:'', guiding the model to produce sentences with fewer than 100 tokens, excluding special symbols.
For the text-to-image generation, we employ Stable Diffusion 3 medium \cite{esser2024scaling}. This model uses three fixed, pre-trained encoders: CLIP L/14 \cite{Radford2021LearningTV}, OpenCLIP bigG/14 \cite{cherti2023reproducible}, and T5-v1.1-XXL \cite{raffel2020exploring}. The entire diffusion model is pre-trained on the CC12M dataset \cite{changpinyo2021conceptual}. Both original and generated images are encoded by DINOv2, which is distilled from ViT-B/14 on the LVD-142M dataset.
For the proposed image caption dataset, human evaluation serves as the validation method for the framework. Each image in the dataset is accompanied by five human-annotated captions, and performance is measured using the average cosine similarity score. 

\vspace{-0.3cm}
\subsection{Human Likert Scale Judgment}
\label{dataset:mscoco_flickr}
\noindent\textbf{Evaluation on Proposed Dataset.}
We aim to test the capability of our method to measure the model’s image caption quality without ground-truth annotation. To ensure the validity of the evaluation results based on our framework—specifically, their alignment with human judgment—we propose a dataset enhanced from the open-domain image caption dataset MSCOCO and an evaluation framework to test the feasibility of our method. In our study, we enhance the existing MSCOCO Caption dataset by incorporating an additional 30,000 human-annotated image-description pairs from Flickr30k. This augmented dataset serves as the basis for evaluating the alignment of our proposed evaluation method with human-annotated image descriptions.

The dataset introduced in this work consists of pairs of images and captions, each accompanied by five distinct human-generated captions. The details of our human evaluation process are outlined below. In Step 1, we use the human-annotated ground truth caption to generate an image through a text-to-image generation model, such as Stable Diffusion or DALL-E. In Step 2, we extract the image features of both the ground truth caption’s corresponding image and the image generated by the text-to-image model. In Step 3, we apply the cosine similarity formula from Section \ref{metric:cosine} to compute the cosine similarity scores between these two sets of image features.
Given that the caption is a human-annotated ground truth description, accurately portraying the corresponding image, we expect the similarity score from Step 3 to be high. Conversely, if a caption inaccurately describes a given image, the cosine similarity score from Step 3 should be low. Consistency between the experimental result and these expectations indicates the effectiveness of the proposed evaluation framework in aligning with human consensus.

\begin{figure}[t!]
\begin{center}
\centering
\scalebox{0.7}{
\includegraphics[width=1.0\linewidth]{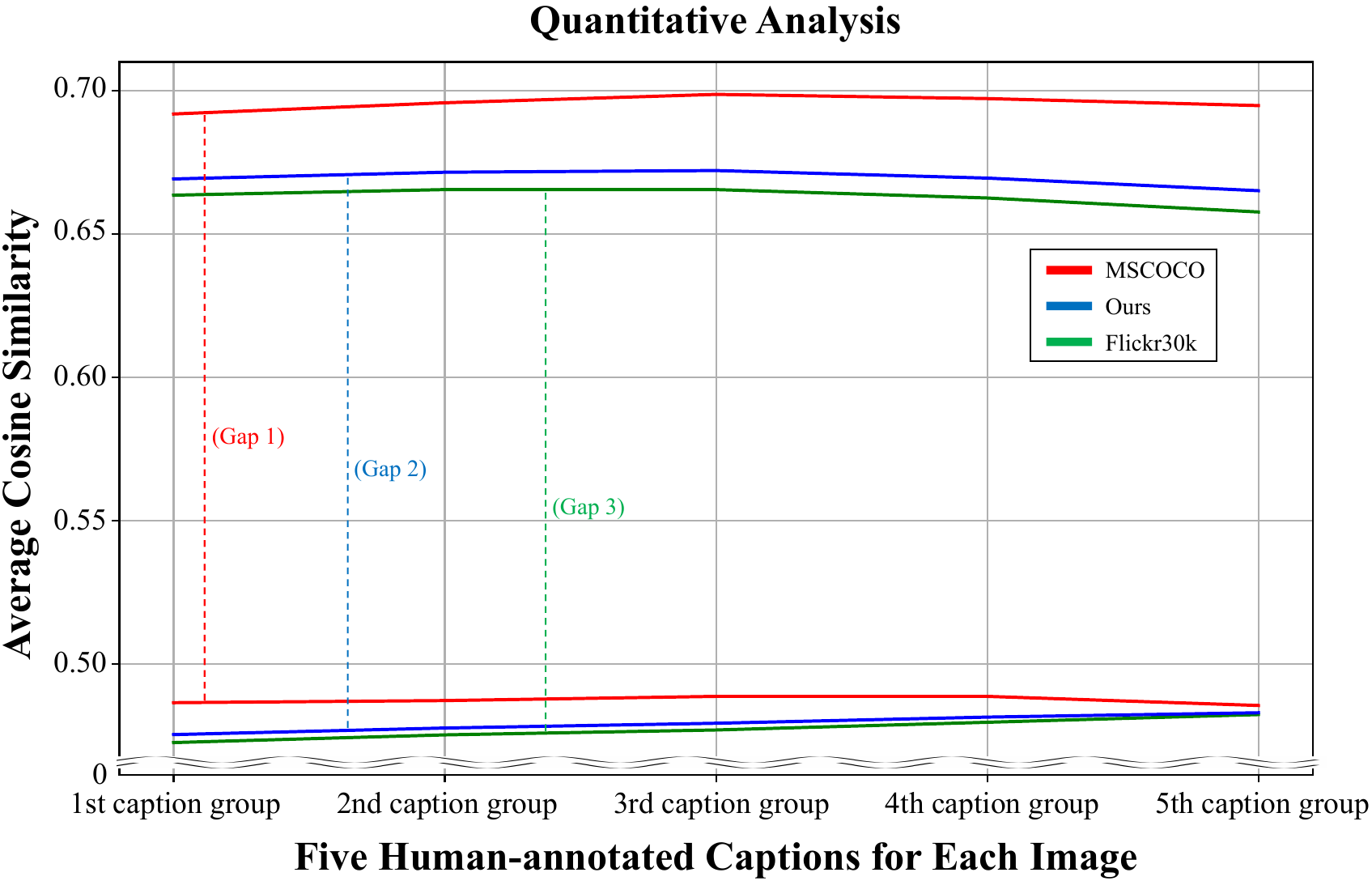}
}
\end{center}
\vspace{-0.7cm}
   \caption{\small Human evaluation results. 
   The top three lines represent scenarios where the provided caption aligns with the correct human-annotated description, while the bottom three lines represent scenarios where the caption is incorrect. ``Gap 1'', ``Gap 2'', and ``Gap 3'' signify the disparities in average cosine similarity scores. 
   }
\vspace{-0.6cm}
\label{fig:figure4}
\end{figure}

The evaluation results depicted in Figure \ref{fig:figure4} reveal notable insights. The blue lines in Figure \ref{fig:figure4} illustrate the impact of the provided captions on the cosine similarity scores. Specifically, when the provided caption matches the correct human-annotated description (upper blue line), the average cosine similarity score reaches approximately 0.67. Conversely, when the caption is incorrect (lower blue line), the average cosine similarity score drops to around 0.47. This discrepancy results in a similarity gap of approximately 0.2. These findings underscore the effectiveness of the proposed evaluation framework, as it closely aligns with human judgment. The robustness of this human evaluation method is attributed to the remarkable text-to-image generation capabilities of modern VLLMs. Widely recognized models such as GPT-4V and Gemini have been extensively acclaimed in various studies and by the broader community \cite{brown2020language,team2023gemini,zhu2024enhancing,huang2024personalized}.

Figure \ref{fig:figure4} reveals consistent trends in the evaluation results across MSCOCO, Flickr30k, and our dataset. Similar patterns are observed in MSCOCO and Flickr30k, where there is a notable decrease in the average cosine similarity when the model-generated image caption differs from the human-annotated ground truth caption. These findings affirm the effectiveness and reliability of the proposed evaluation framework for assessing image captioning models.
\newline

\noindent\textbf{Flickr8K-Expert and Flickr8K-CF \cite{hodosh2013framing}.}
The Flickr8K dataset consists of three main components: images, paired captions (or captions from other images), and human judgment to evaluate the correctness of the captions. Flickr8K-Expert contains $5,664$ image-caption pairs, each annotated by three human experts who provide judgment scores ranging from $1$ to $4$. A score of $1$ indicates that the caption does not describe the image at all, while a score of $4$ indicates that the caption accurately describes the image. Following the CLIPScore setting, we flatten all human judgment scores into a list of $16,992$ 
$(5,664 \times 3)$ data samples. Flickr-CF \cite{hodosh2013framing} is another human-annotated dataset with 145K binary quality judgments collected from CrowdFlower for 48K image-caption pairs and 1K unique images. Each pair has at least three binary human judgments, and we use the percentage of “yes” responses as the judgment score. For both Flickr-Expert and Flickr-CF judgments, we use Kendall’s $\tau$ coefficient \cite{kendall1938new} to evaluate the correlation between the metric and human judgment, with $\tau_{c}$ and $\tau_{b}$ used for Flickr8K-Expert and Flickr8K-CF, respectively.

The experimental results for Flickr8K-Expert are shown in Table \ref{tab:expert-result}, and for Flickr8K-CF in Table \ref{tab:cf-result}. Our method achieves higher correlations with human judgments than CLIP-S (without references) and other previous methods that rely on references. Additionally, our method is only slightly inferior to the best reference-based method RefCLIP-S in the Flickr8K-CF dataset and outperforms it in the Flickr8K-Expert dataset. This provides strong evidence that our method can produce judgments similar to human evaluations for image captions without ground truth captions.
Furthermore, unlike the CLIPScore method, which uses cross-modality (text-to-image) comparison, our method maintains the comparison within the same modality. The image generated from the caption can reconstruct semantic information into spatial information. This comparison within the spatial domain is more contrastive and improves image caption evaluation quality while reducing hallucination. 

\begin{minipage}{\textwidth}
 \begin{minipage}[t]{0.45\textwidth}
  \centering
     \makeatletter\def\@captype{table}\makeatother\caption{Correlations with human judgment for the Flickr8K-Expert.}
     \vspace{+0.1cm}
     \scalebox{0.75}{
       \begin{tabular}{lll}
            \hline
                              & & $\tau_{c}$   \\ \hline
            BLEU-1~\cite{papineni2002bleu}            & & 32.3  \\
            BLEU-4~\cite{papineni2002bleu}            & & 30.8  \\
            ROUGE-L~\cite{lin2004rouge}           & & 32.3  \\
            BERT-S (RoBERTa-F) & & 39.2  \\
            METEOR~\cite{banerjee2005meteor}            & & 41.8  \\
            CIDEr~\cite{vedantam2015cider}             & & 43.9  \\
            SPICE~\cite{anderson2016spice}             & & 44.9  \\
            LEIC($\tau_{b}$)~\cite{cui2018learning}  & & 46.9  \\
            BERT-S++~\cite{yi2020improving}          & & 46.7  \\
            TIGEr~\cite{jiang2019tiger}             & & 49.3  \\
            NUBIA~\cite{kane2020nubia}             & & 49.5  \\
            ViLBERTScore-F~\cite{lee2020vilbertscore}   & & 50.1  \\
            CLIP-S~\cite{hessel2021clipscore}            & & 51.2  \\
            RefCLIP-S~\cite{hessel2021clipscore}         & & 53.0  \\ \hline
            \textbf{Image2Text2Image (Ours)}              & & \textbf{53.5} \\ \hline
            \end{tabular}  
        }
    \label{tab:expert-result}
  \end{minipage} 
  \hspace{0.015\textwidth} 
  \begin{minipage}[t]{0.45\textwidth}
   \centering
        \makeatletter\def\@captype{table}\makeatother\caption{Correlation with human judgment for Flickr8k-CF, a version of Flickr-8k dataset annotated through crowdsourcing.}
        \vspace{+0.1cm}
            \scalebox{0.75}{
            \begin{tabular}{ll}
                \hline
                                   & $\tau_{b}$   \\ \hline
                BLEU-4~\cite{papineni2002bleu}              & 16.9  \\
                ROUGE-L~\cite{lin2004rouge}              & 19.9  \\
                BERT-S (RoBERTa-F) & 22.8  \\
                METEOR~\cite{banerjee2005meteor}              & 22.2  \\
                CIDEr~\cite{vedantam2015cider}               & 24.6  \\
                SPICE~\cite{anderson2016spice}               & 24.4  \\
                LEIC~\cite{cui2018learning}                & 29.5  \\
                CLIP-S~\cite{hessel2021clipscore}              & 34.4  \\
                RefCLIP-S~\cite{hessel2021clipscore}           & \textbf{36.4}  \\ \hline
                \textbf{Image2Text2Image (Ours)}               & 35.2 \\ \hline
            \end{tabular}
            }
        \label{tab:cf-result}
   \end{minipage}
\end{minipage}

\subsection{Evaluation on the Hallucination Sensitivity}
\noindent\textbf{Hallicination Detection on Foil.} In the context of image captioning for VLMs, hallucinations can manifest as responses containing incorrect references or descriptions of the input image~\cite{li2023evaluating}. To mitigate these hallucinations and enhance the reliability and accuracy of image captioning models in real-life use cases, a pragmatic and efficient detection system is required. We first test our method on the FOIL dataset \cite{shekhar-etal-2017-foil} to verify its ability to detect potential misdescriptions in image captions. FOIL reformulates the COCO image caption dataset by swapping a single noun phrase, e.g., substituting “bike” for “motorcycle”. Following previous work, we select sample pairs with FOIL words and compute the accuracy of each evaluation metric in assigning a higher score to the true candidate versus the FOIL. Table \ref{tab:foil_result} shows the results. The results reveal that when the metrics are exposed to more references, the evaluation improves significantly. This is likely because the reference captions contain the original words, leading to better measurement scores. Our method outperforms all reference-free metrics, including CLIP-S, but is slightly inferior to the best reference-based model. This demonstrates its outstanding ability to detect obvious incorrect information within image descriptions without the assistance of references.

\begin{figure}[t!]
\begin{center}
\centering
\scalebox{0.8}{
\includegraphics[width=1.0\linewidth]{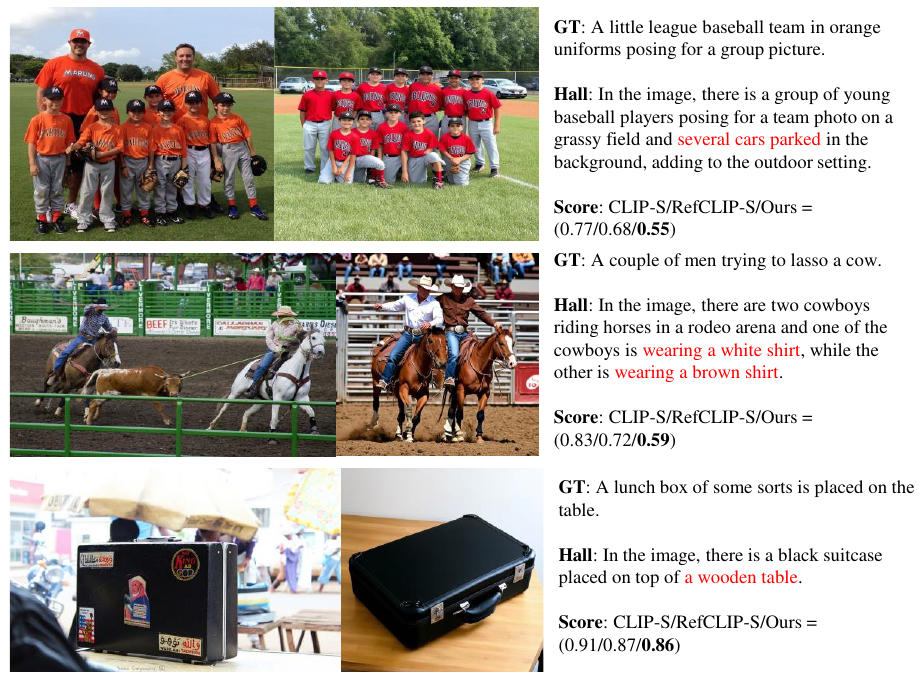}
}
\end{center}
\vspace{-0.7cm}
   \caption{\small The visualization of the hallucination detection. \textbf{GT} and \textbf{Hall} denote ground truth and hallucinated captions, respectively. The hallucinated descriptions are highlighted in red. In each row, the first image represents the original image, and the second image represents the image generated from the hallucinated caption. For each caption, we report the scores of CLIP-S \cite{hessel2021clipscore}, RefCLIP-S \cite{hessel2021clipscore}, and our method. The first two rows show the corrected examples, while the third row shows an example that all three methods failed to detect. A smaller score is better for detecting the hallucinated captions.
}
\vspace{-0.4cm}
\label{fig:figure5}
\end{figure}

\noindent\textbf{Hallicination Detection on M-Hallucination.} The FOIL dataset is designed to have only one mismatched word between the image caption and the original image. However, the FOIL setting cannot simulate the hallucinations that occur in real-world applications. Hallucination text often contains multiple and more complex descriptions of non-existent objects. To assess this aspect, we use the M-HalDetect dataset \cite{gunjal2024detecting}, which includes fine-grained annotations at a sub-sentence level over detailed image descriptions. This dataset consists of image descriptions generated by the recent VLLM InstructBLIP, prompted by randomly selected questions from a pool of instructions for describing an image. For hallucination annotation, each sentence of a generated image description is manually labeled as either “Accurate” or “Inaccurate”. According to the original work setting, we conduct sentence-level binary hallucination classification. Following the previous experiment, we compute the accuracy of each evaluation metric in assigning a higher score to the ground truth image caption compared to the hallucinated parts of the caption sentences generated by a VLLM-based image captioning model. As shown in Table \ref{tab:hall_result}, our method outperformed all baseline models, including the best reference-free model CLIP-S, and the best reference-based model RefCLIP. These results indicate that image caption hallucinations from real-world applications pose a challenge to existing evaluation methods. Our method is capable of effectively evaluating under this scenario without exposure to reference annotations. We can observe from Figure \ref{fig:figure5} that the baseline models like CLIP-s~\cite{hessel2021clipscore} and RefCLIP-s mostly focused on the correct part of the description but ignored the hallucinated part. Our method can reconstruct both correct and hallucinated parts of the description and makes more effective single-modal comparisons with the original image. while some ambiguous object feature descriptions such as whether the ``table'' is wooden can still confuse our method.  

\begin{minipage}{\textwidth}
 \begin{minipage}[t]{0.45\textwidth}
  \centering
     \makeatletter\def\@captype{table}\makeatother\caption{Pairwise accuracy results on the FOIL hallucination detection. The baseline models use either one or four references.}
    \vspace{+0.1cm}
     \scalebox{0.75}{
       \begin{tabular}{lll}
            \hline
                      & 1-ref & 4-ref \\ \hline
            length    & 50.2  & 50.2  \\
            BLEU-4~\cite{papineni2002bleu}     & 66.5  & 82.6  \\
            BERT-S    & 88.6  & 92.1  \\
            METEOR~\cite{banerjee2005meteor}      & 78.8  & 85.4  \\
            CIDEr~\cite{vedantam2015cider}     & 82.5  & 90.6  \\
            SPICE~\cite{anderson2016spice}       & 75.5  & 86.1  \\
            CLIP-S~\cite{hessel2021clipscore}     & 87.2  & 87.2  \\
            RefCLIP-S~\cite{hessel2021clipscore}  & \textbf{91.0}  & \textbf{92.6}  \\ \hline
            \textbf{Image2Text2Image (Ours)}      & 87.86 & 87.86  \\ \hline
            \end{tabular}   
        }
    \label{tab:foil_result}
  \end{minipage} 
  \hspace{0.015\textwidth} 
  \begin{minipage}[t]{0.45\textwidth}
   \centering
        \makeatletter\def\@captype{table}\makeatother\caption{Pairwise accuracy results on the M-HalDetection. The baseline models use one reference because this dataset lacks multiple references.}
        \vspace{+0.1cm}
        \scalebox{0.75}{
         \begin{tabular}{ll}
            \hline
                      & 1-ref  \\ \hline
            length    & 15.3 \\
            BLEU-1~\cite{papineni2002bleu}     & 20.1 \\
            BERT-S    & 34.8 \\
            METEOR~\cite{banerjee2005meteor}     & 28.4 \\
            CIDEr~\cite{vedantam2015cider}     & 32.3 \\
            SPICE~\cite{anderson2016spice}      & 23.6 \\
            CLIP-S~\cite{hessel2021clipscore}    & 35.2 \\
            RefCLIP-S~\cite{hessel2021clipscore} & 38.5 \\ \hline
            \textbf{Image2Text2Image (Ours)}      & \textbf{57.3} \\ \hline
            \end{tabular}
        }
        \label{tab:hall_result}
   \end{minipage}
\end{minipage}

\section{Conclusion}
\vspace{-0.2cm}
In this study, we have introduced a novel framework called Image2Text2Image for evaluating automatically generated image descriptions, aiming to overcome the limitations of existing evaluation metrics like BLEU, ROUGE, METEOR, and CIDEr. Our framework leverages advancements in text-to-image generation models such as Stable Diffusion or DALL-E to utilize image descriptions generated by an image captioning model for creating corresponding images. By quantifying the cosine similarity between the representation of the original input image in the image captioning model and the representation of the generated image, we can effectively assess the image captioning model's performance without relying on human-annotated reference captions.
Through extensive experiments on established datasets like Flickr8k-Expert, Flickr8k-CF, and our proposed dataset, we have demonstrated the effectiveness of the proposed evaluation framework. Our experimental results suggest that the proposed framework's performance closely correlates with human judgment, offering a valuable method for evaluating the effectiveness of image captioning models. Additionally, human evaluations conducted on our introduced dataset validate the framework's efficacy in capturing various aspects such as grammaticality, coverage, correctness, and truthfulness in automatically generated image descriptions.
Moving forward, the proposed framework presents new opportunities for evaluating image captioning models, offering a more efficient and reliable alternative to traditional human evaluations and existing automated evaluation metrics. 
It is designed to complement, rather than replace, human judgment. In summary, our work contributes to the ongoing development of robust evaluation frameworks for image captioning models, bridging the gap between automated metrics and human judgment, and driving advancements in this field.

\bibliographystyle{splncs04}
\bibliography{reference}
%




\end{document}